\RequirePackage{fix-cm}
\documentclass[twocolumn]{svjour3}          
\smartqed  
\usepackage{hyperref}
\usepackage{graphicx}
\usepackage{times}
\usepackage{comment}
\usepackage{latexsym} 
\usepackage{graphics}
\usepackage{graphicx}
\usepackage{microtype}
\usepackage{caption, subcaption} 
\journalname{Applied Intelligence}
\begin{document}

\title{Do Images really do the Talking? 
}
\subtitle{Analysing the significance of Images in Tamil Troll meme classification}

\titlerunning{Do Images really do the Talking}        

\author{Siddhanth U Hegde\and
        Adeep Hande \and
        Ruba Priyadharshini \and
        Sajeetha Thavareesan \and
        Ratnasingam Sakuntharaj \and
        Sathiyaraj Thangasamy\and
        B Bharathi\and
        Bharathi Raja Chakravarthi
}

\authorrunning{Hegde et al.} 

\institute{Siddhanth U Hegde \at
              University Visvesvaraya College of Engineering, Bangalore University\\ 
              \emph{siddhanthhegde227@gmail.com}         
           \and
          Adeep Hande \at
             Indian Institute of Information Technology Tiruchirappalli, Tamil Nadu, India \\
              \emph{adeeph18c@iiitt.ac.in}  
            \and
            Ruba Priyadharshini \at 
            ULTRA Arts and Science College, Madurai, Tamil Nadu, India\\
            \emph{rubapriyadharshini.a@gmail.com}
            \and 
             Sajeetha Thavareesan, Ratnasingam Sakuntharaj \at
            Eastern University, Sri Lanka\\
            \emph{\{sajeethas,sakuntharaj\}@esn.ac.lk} 
            \and
            Sathiyaraj Thangasamy \at
            Sri Krishna Adithya College of Arts and Science, Coimbatore, Tamil Nadu, India. \\
            \emph{sathiyarajt@skacas.ac.in}
            \and
            B Bharathi\at
             SSN College of Engineering, Tamil Nadu, India. \\
             \emph{bharathib@ssn.edu.in}
             \and
            Bharathi Raja Chakravarthi* \at
            Insight SFI Research Centre for Data Analytics, National University of Ireland Galway, Galway, Ireland\\
            \emph{bharathi.raja@insight-centre.org}
}

\date{Received: date / Accepted: date}

\maketitle

\begin{abstract}
A meme is an part of media created to share an opinion or emotion across the internet. Due to its popularity, memes have become the new forms of communication on social media. However, due to its nature, they are being used in harmful ways such as trolling and cyberbullying progressively. Various data modelling methods create different possibilities in feature extraction and turning them into beneficial information. The variety of modalities included in data plays a significant part in predicting the results. We try to explore the significance of visual features of images in classifying memes. Memes are a blend of both image and text, where the text is embedded into the image. We try to incorporate the memes as troll and non-trolling memes based on the images and the text on them. However, the images are to be analysed and combined with the text to increase performance. Our work illustrates different textual analysis methods and contrasting multimodal methods ranging from simple merging to cross attention to utilising both worlds' - best visual and textual features. The fine-tuned cross-lingual language model, XLM, performed the best in textual analysis, and the multimodal transformer performs the best in multimodal analysis.
\end{abstract}

\section{Introduction}
\label{intro}
Easier access to the internet has assisted the user-base of social media platforms to communicate and express their perspective about anything without any censorship \cite{ghanghor-etal-2021-iiitk}. Over the past decade, memes have been used as a medium of communication, expressing users' intentions about particular things. Memes occur in several forms, such as image, text, and video. They are often used to spread knowledge, emotions, ideas, and talents. Due to its vast popularity, the social media handles of government agencies and industry professionals use memes to promote awareness programs, advertise their products, ideas, and so on \cite{mishra-saumya-2021-iiit}.
However, a meme that has been perceived as funny by one could probably be perceived as offensive by others \cite{chen2012creation}. The main characteristic of a meme is that it can be changed, recreated, and often be taken out of context to be used for sarcastic perspectives \cite{grundlingh2018memes,8354676}. However, several memes are created to demean people based on their gender, sexual orientation, religious beliefs, or any other opinions, often regarded as trolling, which could cause distress in the online community \cite{suryawanshi-chakravarthi-2021-findings}.

The most popular memes available on social media applications are image with text (IWT) memes. IWT memes \cite{Du_Masood_Joseph_2020} make it harder to decode its intention or any other characteristics \cite{avvaru-vobilisetty-2020-bert,Nave2018TalkingIP}. The comprehensive analysis of IWT could elucidate the socio-political and societal factors, their implications on cultures and the values promoted by them \cite{Milner2013FCJ156HT}. One of the alternatives to manually moderating memes on social media platforms is to devise automated systems that would classify any meme as to be trolling or not.

IWT memes are used in daily conversations and across all social media platforms. In multilingual countries such as India, where languages represent cultures, a meme could represent a culture and henceforth could be used to troll certain cultures and lifestyles. To analyse the memes, we develop a multimodal approach to classify images and texts by employing pretrained language models for texts while using pretrained vision models to classify the images in IWT memes. 

In a meme, attributes like sarcasm, satire, and irony are brought out usually in captions, along with the occasional referencing to the images. Hence, we try several multimodal approaches. We then compare the performances of these models with the performances of transformer-based models for code-mixed Tamil texts. Theoretically, the visual features are supposed to elevate the model's overall performance when used with the textual features during multimodal analysis. Despite using transfer learning, while using datasets of smaller sizes ( \~ 2500 Images), it has been observed that the visual features did not directly influence the overall performance in identifying the memes that troll people. This drop in the performance maybe related to the nature of the dataset, as the dataset was obtained by scraping memes from the Internet, which had texts written over the images, and automated transformation of images did not remove all of the texts. The major problems we believe, that are faced by the models are the lack of data.

The rest of the paper is organised as follows. Section \ref{Section 2} shows previous work on analysing memes using unimodal and multimodal approaches. Section \ref{Section 3} consists of a detailed description of the datasets for our purposes. Section \ref{Section 4} talks about the proposed model with Sections \ref{Section 5} and \ref{Section 6} using unimodal  and multimodal approaches respectively. We describe the experimental results and error analysis in Sections \ref{Section 7} and Section \ref{Section 8} respectively, and conclude our work and discuss potential directions for our future work in Section \ref{Section 9}.
\section{Related Work}
\label{Section 2}
Analysing the image part of the meme falls under the domain of computer vision. Image classification was a pretty difficult task back then (before the 1960s). However, now several methods and techniques have been introduced to mimic the human visual system. Breakthrough in this field was by introducing the convolutional neural networks \cite{article4} which gave marvellous results. Numerous applications have developed such as object detection \cite{9079525}, image segmentation \cite{inproceedings2}, biomedical applications \cite{9118916} and many more. Recent trends include generating high quality images \cite{goodfellow2014generative} and image translations \cite{isola2017image}. Transfer learning has made training deep learning models easier by transferring the weights of a bigger task to smaller downstream tasks. Studies also show that transfer learning results and fine-tuning models are significantly better than training models from scratch \cite{10.3844/jcssp.2021.44.54}. Transformers for images significantly improved computer vision through their attention mechanism on images \cite{dosovitskiy2020image}. These consider images as a series of patches and try to attain the conventional attention mechanism on them. Many more researchers followed this in their work of computer vision for numerous other tasks. 

As we try to analyse text in our task, we look into Natural Language Processing(NLP). The final goal of such analysing texts is to perform repetitive tasks like summarization \cite{textsummar}, language translation\cite{nmt}, spam classification\cite{spam} and many more. NLP is performed by preprocessing texts and converting them into meaningful numbers/vectors. Bag of words \cite{9079815} methodology was introduced to simplify and retrieve information from the text. The point was to give numbers and convert them into n-grams. For example, a bi-gram would be a phrase with two words. Different words will be added to the next n-gram, and the shortage of words is handled by padding the n-grams. This extra data caused input data to have more features, thus decreasing performance due to the curse of dimensionality. It also assumed words are independent of each other. So every word was represented in a vector space \cite{mikolov2013efficient},  providing a meaning to the words and thus helped in sequential models. Nevertheless, the breakthrough in NLP was with the introduction of attention mechanism \cite{vaswani2017attention} on the sentences. The model performed immensely well on almost all NLP tasks.
Additionally, such models could be fine-tuned for downstream tasks. The models had millions of parameters and required higher computational power. In our work, we try to accommodate transformers for text analysis. 

Multimodal analysis refers to interpreting and making sense of qualitative data in projects that mix verbal and nonverbal forms of information\footnote{\url{https://methods.sagepub.com/foundations/multimodal-analysis}}. It is the process of extraction of information retrieval by accepting the joint representation of all the modalities used in the system. As it combines the properties of various aspects, it might seem as adding more valuable data might improve results, but it is not always true that multimodal models will always boost performance. It can add constant noise and other losses as well \cite{8715409}. This journal focuses on the main ideas behind how image and text data can be combined and the requirement of multimodal analysis on Tamil troll meme classification. A multimodal sentiment analysis system was developed \cite{Hu_2018} by devising a deep neural network that combines both visual analysis and text analysis to predict the emotional state of the user by using Tumblr posts. This analysis illustrates the importance of analysing emotions hidden behind the uploads of the user based on their day-to-day life cycle. Such sentiments can be found out and thus could be discarded or kept by the admin based on specific criteria decided for the application. .When it comes to multimodal analysis on Dravidian languages, several systems were submitted as a part of a shared task on classifying Tamil Troll memes \cite{suryawanshi-chakravarthi-2021-findings}. A total of ten systems were submitted to the shared task, with most of the researchers treating it as a multimodal task, trying to mutually address the textual and visual features by devising deep neural networks that jointly learns from the features. 
     
Our contribution is an extension of our previous work \cite{u-hegde-etal-2021-uvce}. This paper highlighted its architecture opting for a complete attention-based architectures for both textual and visual analysis by not using Convolutional Neural Networks(CNN) and Recurrent Neural Networks(RNN). It was a pure transformer-transformer architecture where both image and text encoders used transformers to extract features. The model scored a perfect F1 score of 1.0 on the train and validation set but scored an F1 score of 0.47 on the test dataset because of the similar issues mentioned in Section 1.  
\captionsetup[sub]{labelformat=simple} 
\renewcommand{\thesubfigure}{(\alph{subfigure})}
\begin{figure}[htpb] 

    \begin{subfigure}[b]{0.9\linewidth}  
        \centering 
        \includegraphics[width=\textwidth]{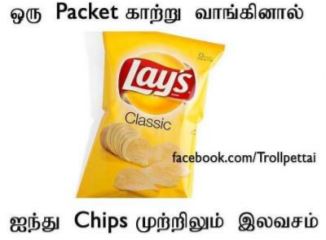} 
        \caption{Example of an image belonging to a troll class} 
        \label{fig:1-1} 
    \end{subfigure} 
    \\ 
    \begin{subfigure}[b]{0.9\linewidth} 
        \centering
        \includegraphics[width=\textwidth]{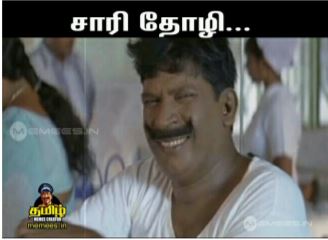} 
        \caption{Example of an image belonging to non-troll class} 
        \label{fig:1-2} 
    \end{subfigure} 
    \caption{Examples of the dataset} 
    \label{fig:1}  
\end{figure}
\section{Dataset}
\label{Section 3}
\begin{figure*}
    \centering
    \includegraphics[width=16cm, height=9cm]{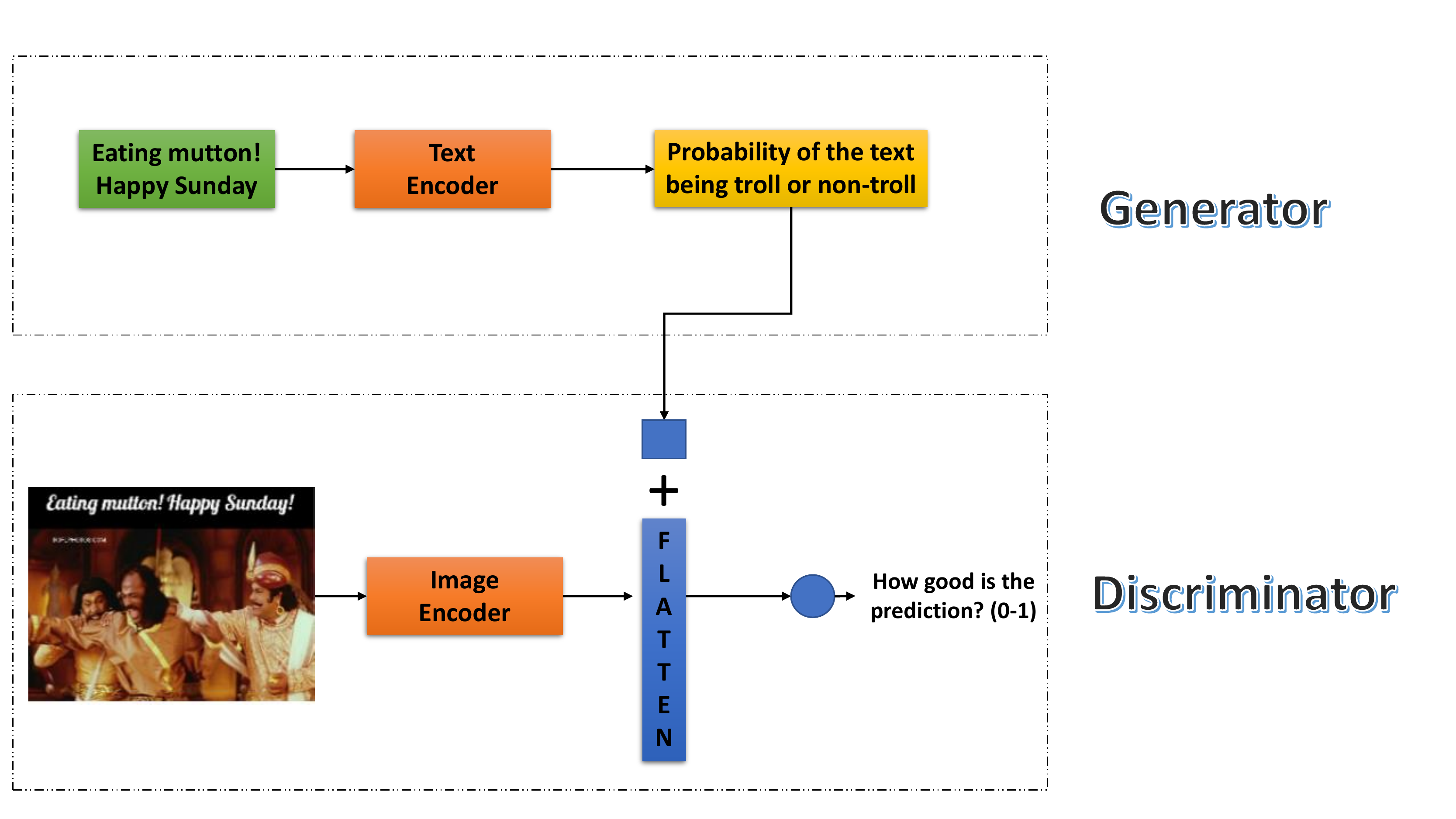}
    \caption{Architecture of Adversarial learning}
    \label{fig:adversarial}
\end{figure*}
We use the troll classification dataset of Tamil Memes \cite{suryawanshi-etal-2020-dataset}. It consists of 2,699 memes, of which most of the images have text embedded within them. We were also provided with captions for all images.  Fig.\ref{fig:1-1} and Fig.\ref{fig:1-2} have the code-mixed Tamil-English captions embedded into the image. We have added these images for easier understanding. Fig.\ref{fig:1-1} tries to troll the pack of chips, stating that \textit{If you buy a packet of air, 5 chips are completely free}, while Fig.\ref{fig:1-2} does not intend to be trolling, only intending to sarcastically be apologetic to \textit{girlfriend}. The images are certain still frame from a movie or TV shows in Tamil languages. The distribution is shown in Table \ref{tab1}. The dataset consists of two classes:
\begin{itemize}
    \item \textbf{Troll:} A troll meme is an implicit image that intends to demean or offend an individual on the internet.
    \item \textbf{Non-Troll:} A meme that does not intend to demean or offend anyone on the internet is non-troll.
\end{itemize}
 
\begin{table}[htbp]\centering
\begin{tabular}{lrrr}
\noalign{\smallskip}\hline
Class &  Train & Validation & Test\\
\noalign{\smallskip}\hline
Troll & 1,154 & 128 & 395\\
Non-Troll & 917 & 101 & 272\\ 
\noalign{\smallskip}\hline
total & 2,071 & 229 & 667 \\
\noalign{\smallskip}\hline
\end{tabular}
\caption{Dataset Distribution}
\label{tab1}
\end{table}
  
\section{Models}
\label{Section 4}
\subsection{Textual Analysis}
\label{Section 5} 
This section discusses the natural language models used for classifying whether a given meme is a troll or a non-troll based on the captions. The Tamil Troll meme dataset \cite{suryawanshi-etal-2020-dataset} comprised of two components, images and captions, both provided separately. We have fine-tuned six pretrained language models. 
\subsubsection{Bidirectional Encoder Representation Transformer (BERT):}
BERT is a language model that pretrains unlabelled data using deep bidirectional representations \cite{devlin-etal-2019-bert}. BERT uses two pretraining strategies, Masked Language Modeling (MLM) and Next Sentence Prediction (NSP). During pretraining, the authors had masked 15\% of the words, which would be predicted later. The function of Next Sentence Prediction is to predict whether a given sentence follows the previous sentence. BERT is pretrained on eleven downstream tasks of NLP. During tokenisation, the tokeniser adds unique tokens, [CLS] and [SEP], at the beginning and end of the sentence, respectively. We then extract the output of the pooled layer ([CLS] token) during fine-tuning to predict the test set. We fine-tune Multilingual BERT, a multilingual language model that was pretrained for the top 104 languages available on Wikipedia dumps. In mBERT \cite{pires-etal-2019-multilingual}, the high-resource languages are downsampled to address the data imbalance during pretraining.

\subsubsection{XLM-RoBERTa:}
XLM-RoBERTa is a multilingual language model that uses self-supervised techniques to improve its overall performance in cross-lingual understanding. XLM-R is pretrained on over 2.5 TB of unlabelled data, focusing more on the low-resourced languages. XLM-R is an improvement over its sister models, RoBERTa and XLM, using BPE (Byte-Pair Encoding) as the preprocessing technique instead of the workpiece tokeniser used in BERT. XLM-Ruses dual-language modelling with Translated Language Modeling (TLM) pretrained over BPE to achieve state-of-the-art results in several downstream tasks, outperforming BERT. XLM-R has three language modelling strategies; \\
(i) \textbf{Masked Language Modeling (MLM):} This language modelling is similar to the approach used in monolingual `Vanilla' BERT.\\
(ii) \textbf{Translated Language Modeling (TLM):} To achieve TLM, every training sample consisted of texts in two languages, with the intention that one model uses the context of one language to predict the tokens of the other language while retaining the same strategy of masking the words randomly.\\
(iii) \textbf{Causal Language Modeling (CLM):} In CLM, a given training sample is trained only based on the existence of previous words while not using any masking strategies.\\

\subsubsection{XLM:}
The XLM model was proposed in cross-lingual language model pretraining 
\cite{dai-etal-2019-transformer}. This model uses a shared vocabulary for different languages. For tokenising the text corpus, Byte-Pair encoding (BPE) was used. Causal Language Modelling (CLM) was designed to maximise the probability of a token \(x_t\) to appear at the \emph{t}th position in a given sequence. Both CLM and MLM perform well on monolingual data.
 \begin{figure*}
    \centering
    \includegraphics[width=16cm, height=9cm]{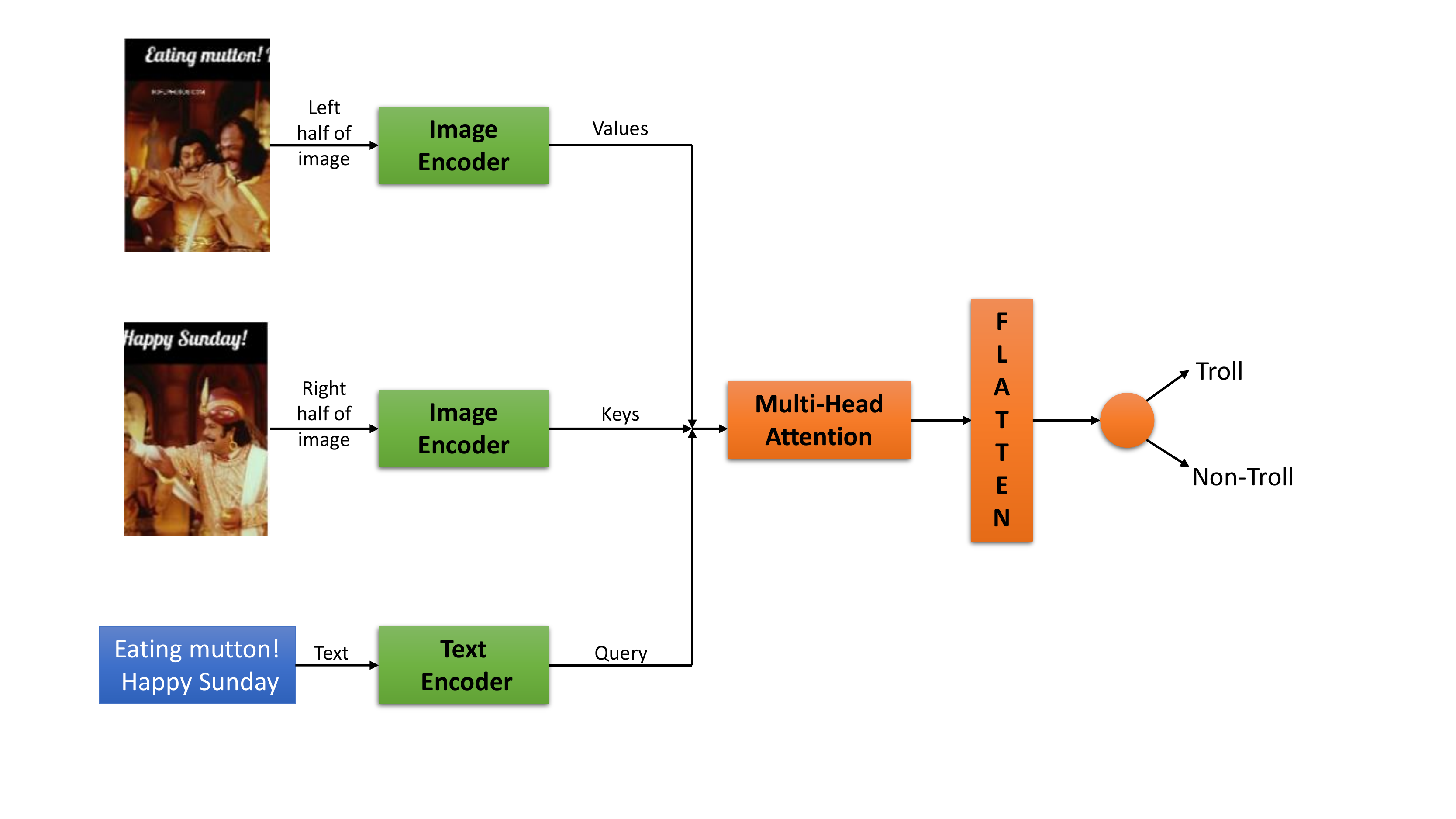}
    \caption{Architecture of EmbracenNet}
    \label{fig:Architecture of EmbracenNet}
\end{figure*}
\subsubsection{Multilingual Representations for Indian languages (MuRIL):}
MuRIL is a language model that focuses on Indian languages, which is not observed in other multilingual models, as the latter are pretrained over hundreds of languages, inherently resulting in the smaller representations of Indian languages \cite{DBLP:journals/corr/abs-2103-10730}. To address the low representations in other multilingual language models, MuRIL was introduced, supporting 17 Indian languages inclusive of English. It follows the architecture of the base model of BERT, only differing the pretraining strategies and data used. Similar to XLM-R, MuRIL uses both supervised and unsupervised language modelling approaches, with the conventional MLM that essentially uses monolingual data for pretraining, while TLM as a supervised learning approach that uses both translated and transliterated documents pair during pretraining. To address data imbalance during pretraining, the low-resource data are upsampled while the high-resourceful data are downsampled to smoothen the data. The model is pretrained from scratch using the Wikipedia\footnote{\url{https://www.tensorflow.org/datasets/catalog/wikipedia}}, Common Crawl\footnote{\url{http://commoncrawl.org/the-data/}}, PMINDIA\footnote{\url{http://lotus.kuee.kyoto-u.ac.jp/WAT/indic-multilingual/index.html}} and Dakshina corpora \cite{roark-etal-2020-processing}.

\subsubsection{TaMillion BERT:}
TaMillion BERT is a monolingual language model that follows the architecture of Efficiently Learning an Encoder that Classifies Token Replacements Accurately(ELECTRA)\cite{clark2020electra} pretrained on 11GB of IndicCorp Tamil\footnote{\url{https://indicnlp.ai4bharat.org/corpora/}} and the Wikipedia dumps\footnote{\url{https://ta.wikipedia.org}} (482 MB) as of October 1, 2020. We use the second version of TaMillion BERT, which has been pretrained on TPU with 224,000 steps. This model significantly outperforms mBERT \cite{pires-etal-2019-multilingual} on classification tasks. ELECTRA is an architecture trained to distinguish the fake tokens from the real tokens, similar to the architecture of GANs.

\subsubsection{Language-agnostic BERT Sentence Embeddings (LABSE):}
LABSE are generated by adapting multilingual BERT. Unlike previous multilingual language models applied to generate English sentence embeddings by fine-tuning pretrained BERT, these models have not been applied to produce multilingual sentence embeddings. Henceforth, the LABSE model combines MLM and TLM pretraining with a translation ranking task using bi-directional dual encoders \cite{DBLP:journals/corr/abs-2007-01852}. LABSE supports 109 languages that use the approach of adopting a pre-trained BERT encoder model to dual encoder model to train the cross-lingual embedding space productively.

All the models used here are really powerful in handling Tamil texts because of their intense pre-training on huge datasets. These transformers would elevate the results in classifying the memes based on their texts. Millions of trained parameters in these transformers can significantly contribute towards our downstream task. Cross Lingual models like XLM and XLM-R was used as the representations of one language influences the predictions of other language. So, these models handle code-mixed and roman script very well.
\begin{figure*}
    \centering
    \includegraphics[width=16cm, height=9cm]{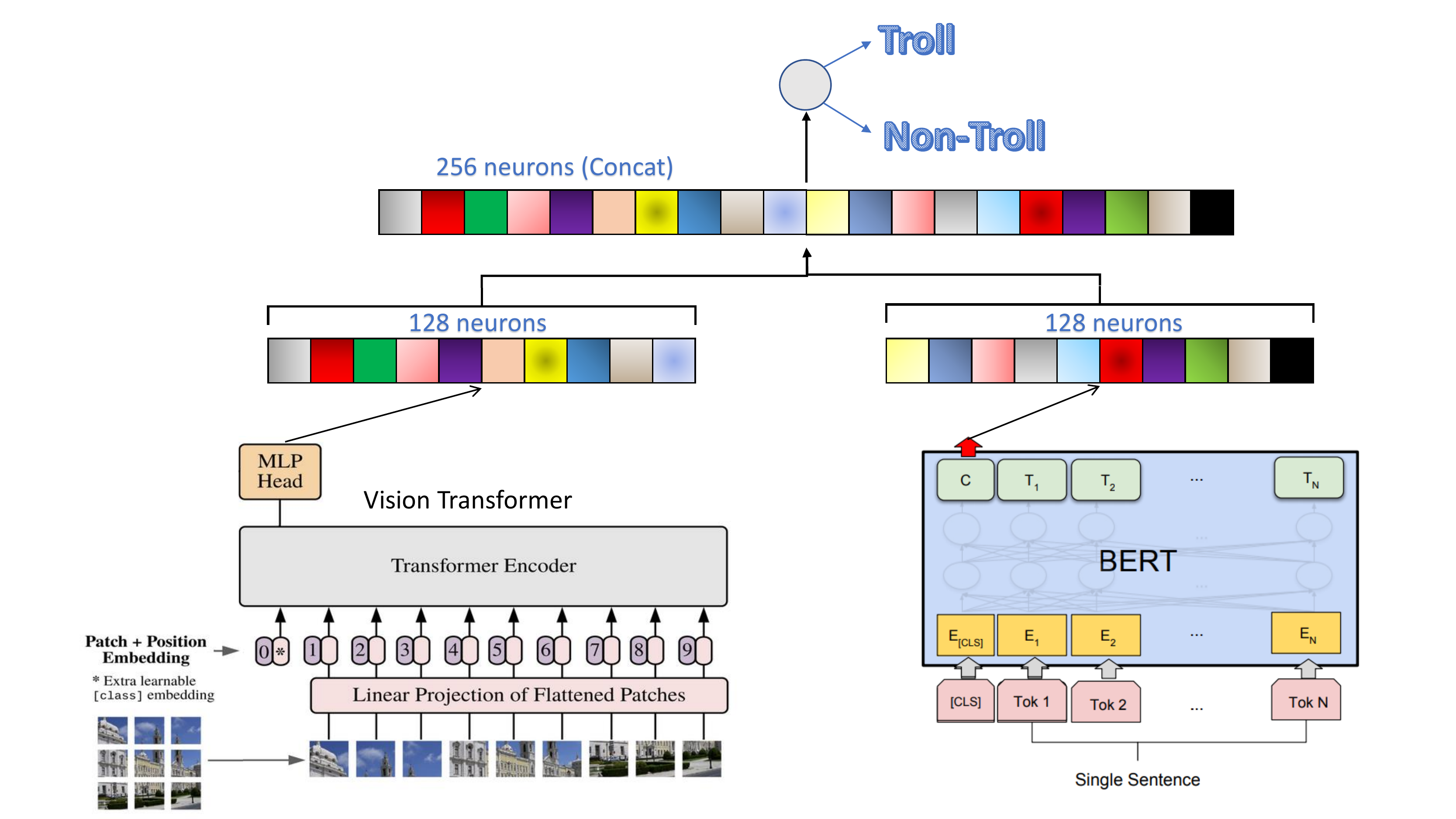}
    \caption{Concatenating textual and visual features \cite{u-hegde-etal-2021-uvce}}
    \label{fig:uvce}
\end{figure*}
\subsection{Multimodal Analysis}
\label{Section 6}
Considering the images, we hope to improve the test results by extracting features and adding them along with them. When the features of text and images are encoded, multiple ways are possible to combine them. We try to focus on the popular architectures used in the multimodal analysis and experiment with the hyperparameters to get the best model out of them. 

\subsubsection{Concatenation}
 
Concatenation is a straightforward approach for merging the features. Here, images are fed into a model trained on the ImageNet dataset \cite{5206848} by fine-tuning it for this task and the embedded texts are fed into a multilingual text model and is fine-tuned to get the features. Then, the outputs of both the sections are merged by stacking one over the other and then getting the predictions for binary classification.
Advanced models try to compete in the ImageNet Large Scale Visual Recognition Challenge (ILSVRC)\cite{ILSVRC15} competition whose objective is to classify millions of images categorised into thousands of classes. Those pretrained models are fine-tuned on other downstream tasks by users, reducing computation power usage and time consumption for training. One of such models is the vision transformer\cite{dosovitskiy2020image}. Keeping the analogy of sentences, instead of 1D token embeddings as input, ViT receives a sequence of flattened 2D patches. If H, W are the height and width of the image and (P, P) is the resolution of each patch, \(N=HW/P^2\) is the effective sequence length for the transformer. Then the patches are projected linearly and then multiplied with an embedding matrix to eventually form patched embeddings. The patches, along with position embeddings, is sent through the transformer. Along with this, a [CLS] token is prepended to determine the class. Here, the number of classes would be the hidden shape to encode the features. Multilingual BERT is used to extract features from the text, and they are also the same token is used to determine the encoded shape. The encoded parts are combined to get a single layer with the shape equal to the encoded image shape + encoded text shape. This encoded text is then carried over to a single output determining the probability of the text being Troll or Non-Troll. The architecture is shown in Fig.\ref{fig:uvce}. 
\begin{figure*}
    \centering
    \includegraphics[width = \linewidth, height=10cm]{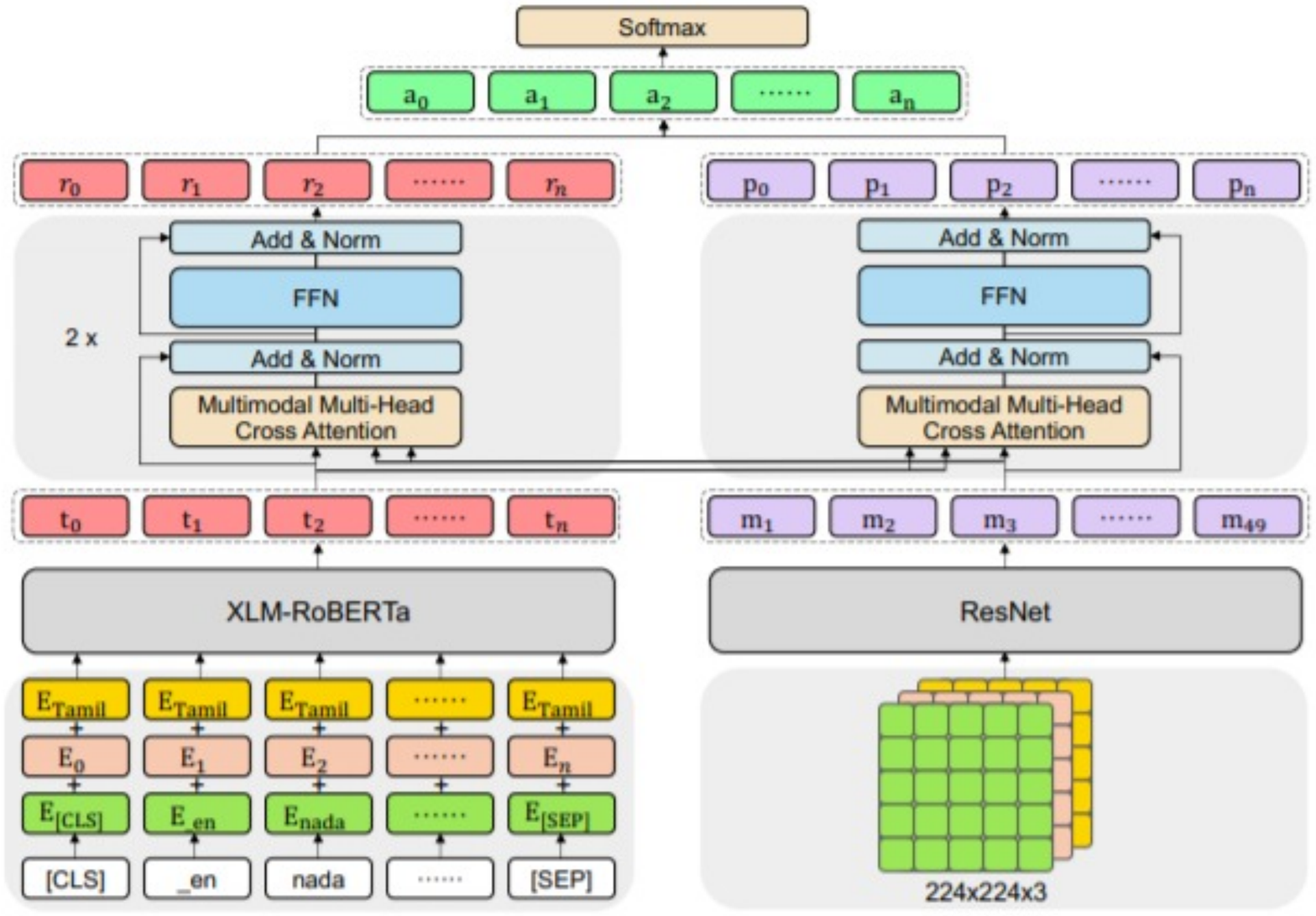}
    \caption{Architecture of Multimodal transformers \cite{li-2021-codewithzichao-dravidianlangtech}}
    \label{fig:xlmr-resnet}
\end{figure*}
\subsubsection{EmbraceNet}
A robust deep learning architecture \cite{choi2019embracenet} used for multimodal classification was used here. Moreover, an extra layer of multi-head attention was used to combine to features. The architecture proposed a solution by feeding halves of the images either cropped vertically or horizontally into two different models instead of feeding an entire image. Cutting the images into two parts creates a high probability of dividing multiple images as memes might contain more than a single image. So, the images were cut vertically in the middle and fed into different imagenet models to extract features. The text was passed through a multilingual transformer as usual. Three outputs are generated from the model, two from the image and one from the text. These three were merged through multi-head attention with encoded text as query and encoded halves of the images as keys and values. The result of this layer was fed into linear layers, and then classification probabilities were generated to classify the memes into a troll or non-troll. Multi-Head attention is a concept of having several workers referred to as heads do their self-attention tasks. It is also known as scaled dot-product attention and is mathematically computed using three vectors from two image encoders and one text encoder, \emph{Query, Key} and \emph{Value} vectors. \emph{Key} and \emph{Value} assume dimensions \(d_{k}\) and \(d_{v}\) respectively. A softmax function is applied on the dot product of queries and keys to compute the weights of the values. In reality, the attention function is computed continuously on a set of queries and then stacked into a matrix Q, being packed into a matrix Q. The \emph{Keys} and \emph{Values} are packed into matrices K and V. The matrix of outputs is computed as follows:
\begin{center}
\begin{equation}
Attention(Q,K,V) =   softmax(\frac{QK^T}{\sqrt{d_k}})V
\end{equation}
\end{center}
This equation represents a single head of self-attention. Similarly, multiple heads are parallel, and each head computes a different set of attention weights and outputs. The architecture of EmbraceNet is shown in Fig.\ref{fig:Architecture of EmbracenNet}.
 
\subsubsection{Multimodal Transformers}
 
This is a concept developed to attain cross attention between two different modalities \cite{li-2021-codewithzichao-dravidianlangtech}. XLM-Roberta \cite{conneau2019unsupervised} was used which was trained on 100 languages Common Crawl dataset as text encoder and ResNet\cite{he2016deep} with 152 layers for image feature extraction followed by linear transformation to match the word embedding size of text encoder. Next is the multihead cross attention which is similar to the conventional multi-head attention described in the previous section. First step was creating attentive word representations for each image. If Q is the query and K is the Key, then the representation R is given by,
\begin{center}
\begin{equation}
A =  LN(Q + Attention(Q,K,K))
\end{equation}
\end{center}
\begin{center}
\begin{equation}
R = LN(A + FFN(A))
\end{equation}
\end{center}
where, LN is a layer normalisation and FFN is feed forward network. Similarly attentive image representations are generated for each word which now takes in K as query and Q as key, and is represented by I,
\begin{center}
\begin{equation}
Z =  LN(K + Attention(K,Q,Q))
\end{equation}
\end{center}
\begin{center}
\begin{equation}
I = LN(Z + FFN(Z))
\end{equation}
\end{center}
Finally, both the representations were concatenated and subjected to average pooling to then get the probability of the class as shown in Fig.\ref{fig:xlmr-resnet}.
 
\begin{table*}[htbp]
    \centering
    \begin{tabular}{l|rrr}
    \noalign{\smallskip}\hline
   Team Name & Precision & Recall & F1-Score\\
    \noalign{\smallskip}\hline
    \textbf{Our Approach} & \textbf{0.60} & 0.58&\textbf{0.57}\\
    Codewithzichao \cite{li-2021-codewithzichao-dravidianlangtech}  &0.57 & \textbf{0.60} & 0.55\\
     IIITK \cite{ghanghor-etal-2021-iiitk}  &0.56 & 0.59  &0.54\\
      NLP@CUET \cite{hossain-etal-2021-nlp} & 0.55  &0.58 & 0.52\\
     SSNCSE NLP \cite{b-a-2021-ssncse-nlp} & 0.58 & 0.60 & 0.50\\
      Simon \cite{que-2021-simon-dravidianlangtech} & 0.53 & 0.58 & 0.49\\
     TrollMeta \cite{j-hs-2021-trollmeta} & 0.45  &0.41  &0.48\\
     UVCE-IIITT \cite{u-hegde-etal-2021-uvce}  &0.60 & 0.60 & 0.46\\ 
      HUB \cite{huang-bai-2021-hub-dravidianlangtech} & 0.50 & 0.54 & 0.40\\
    IIITDWD \cite{mishra-saumya-2021-iiit} & 0.52 & 0.59 & 0.30\\
     \noalign{\smallskip}\hline
    \end{tabular}
    \caption{Comparisons of the existing models developed for the Tamil Troll meme dataset, as a part of the shared task \cite{suryawanshi-chakravarthi-2021-findings}}
    \label{tab:2}
\end{table*}
\begin{table*}[htpb]
    \centering
    \begin{tabular}{l|rrr|rrr|rrrr}
        \hline
        Model & \multicolumn{3}{c|}{Troll} & \multicolumn{3}{c|}{Not-troll} & \multicolumn{4}{c}{Overall}\\
        \noalign{\smallskip}\hline
         & P & R & F1 & P & R & F1 & Acc & W\(_{avg}\)(P) & W\(_{avg}\)(R) & W\(_{avg}\)(F1)\\
        \hline
        mBERT &0.60 &\textbf{0.95}  &0.73 &0.55 &0.08 &0.14 &0.59 &0.58 &0.59 &0.49\\ 
        DistilmBERT &0.60  &0.88 &0.72 &0.51 &0.17 &0.26 &0.59 &0.56 & 0.59 &0.53\\
        XLM-R\(\_{base}\) &0.60 &0.90 &0.72 &0.52 &0.15 &0.23 &0.59 &0.57 &0.59 &0.52  \\
        XLM &\textbf{0.62} &0.81 &0.70 &0.52 &0.29 &0.37 &0.60 &0.58 &0.60 &\textbf{0.57}\\
        MuRIL &0.60 &0.85 &0.71 &0.49 &0.19 &0.28 &0.58 &0.56 &0.58 &0.53\\
        TamillionBERT &0.42 &0.36 &0.39 &0.59 &0.65 &0.62 &0.53 &0.52 &0.53 & 0.52\\
        LABSE &0.50 &0.20 &0.29 &0.61 &\textbf{0.85} &\textbf{0.71} &0.59 &0.56 &0.59 &0.54\\ 
        \noalign{\smallskip} \hline
        Concatenation &0.60 &0.98 &\textbf{0.74} &\textbf{0.60} &0.03 &0.06 &0.60 &0.60 &0.60 &0.47\\ 
        Multimodal Transformers &0.61  &0.75 &0.67 &0.46 &0.31 &0.37 &\textbf{0.61}  &0.57 &0.60 &\textbf{0.55}\\
        EmbraceNet &0.60 &0.95 &0.74 &0.56 &0.09 &0.15 &0.60 &0.58 &0.60 &0.50  \\
        Adversarial Model &0.59  &0.96 &0.73 &0.39 &0.03 &0.06 &0.58 &0.51 &0.58  &0.46\\
        \noalign{\smallskip}\hline
        
    \end{tabular}
    \caption{Results of Textual architectures and Multimodal architectures considering both images and texts}
    \label{tab:3}
\end{table*}
\label{tab: 3}
\subsubsection{Adversarial Learning}
This type of concept was initially built by creating two models, who are trained simultaneously a generative model G that captures the data distribution and a discriminative model D that estimates the probability that a sample came from the training data rather than G\cite{goodfellow2014generative}. Adversarial learning is a two-player minimax game with one player trying to minimise its loss, consequently increasing the other player's loss and vice versa. The idea originated with an aim towards image generation, and this process can be applied in the multimodal analysis to check how well the texts are being classified. We build a generative model G for Tamil text classification using a transformer. This generator outputs a value indicating the probability of the text being troll or non-troll, as shown in Fig.\ref{fig:adversarial}.

\textbf{G} consists of multilingual BERT for encoding, and the model is followed by linear layers and ReLU activation functions, resulting in a single probability with sigmoid activation indicating the class of the text. D consists of an ImageNet such as ResNet\cite{he2016deep} model for encoding the images and is followed by linear layers with a ReLU activation function. The connection between the two models is that the probability of G is concatenated with the encoded image features and then subjected to output a single probability which indicates how good the prediction of G was. The two models work against each other, learning the parameters. The test set was fed into the only G to predict the classes.


\section{Results and Analysis}
\label{Section 7}
\begin{figure}
    \centering
    \includegraphics[width=\linewidth, height=6cm]{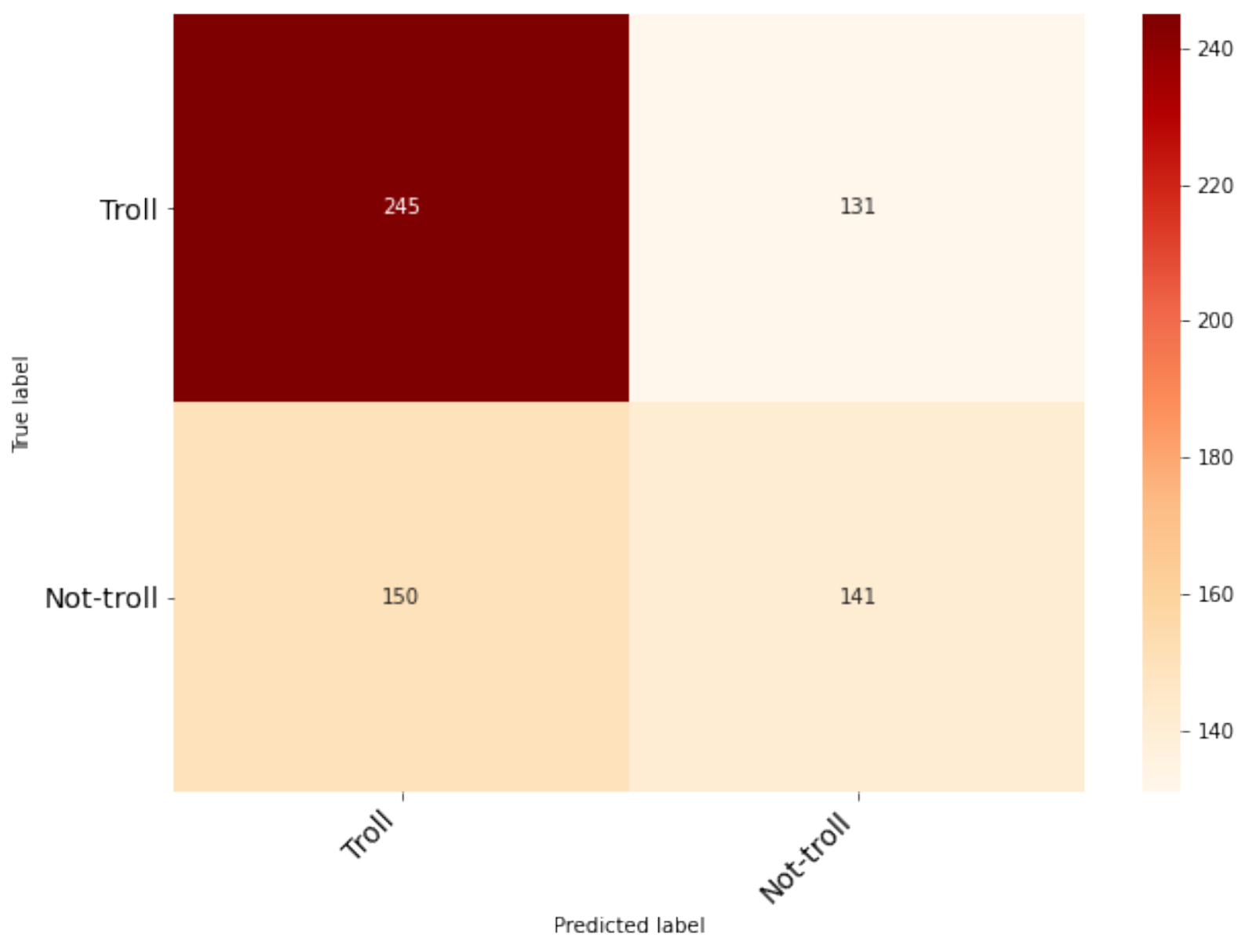}
    \caption{Heatmap of Confusion matrix for the best performing model}
    \label{fig:6}
\end{figure}

All the experiments were conducted on Google Colaboratory\footnote{\url{https://colab.research.google.com/}} accelerated by GPU. For the textual analysis, the transformers were fine-tuned with an optimal learning rate of 2e-5 with Adam optimiser and warmup with the scheduler. The batch size used was chosen among 16,32, and 64 sets according to the computational power. The results of the test set can be observed in Table \ref{tab: 3}. Among all the transformers, XLM scored the highest with a weighted F1 score of 0.57 and next to it is the Language agnostic BERT with a weighted F1 score of 0.54. We have listed the heatmap of the confusion matrix of XLM in Fig.\ref{fig:6}. In other textual models, we observe that the performance of the models is quite similar to each other. 

Interestingly, we see that the Adversarial Model and Concatenation are the two multimodal approaches that score the least of all. As transformers are huge models with millions of parameters, it is difficult to train them perfectly for small datasets. The number of sentences used was only in thousands, and it completely overfits both the training and validation set with more than 0.9. However, the test set results are less due to this high variance generated by the models. 

In comparison to the submitted systems for the Tamil Troll meme detection shared task, and we observe that the text-based unimodal models perform better than multimodal models, as we achieve the best-weighted F1-Score among all the teams. The scores of all the teams are listed in Table \ref{tab:2}. Most of the teams submitted unimodal systems that relied on pretrained natural language models to capture its textual features \cite{ghanghor-etal-2021-iiitk}, while some teams resorted to multimodal approaches that performed worse than the textual models \cite{li-2021-codewithzichao-dravidianlangtech,huang-bai-2021-hub-dravidianlangtech,hossain-etal-2021-nlp}. 
\begin{figure}
    \centering
    \includegraphics[width=\linewidth, height=6cm]{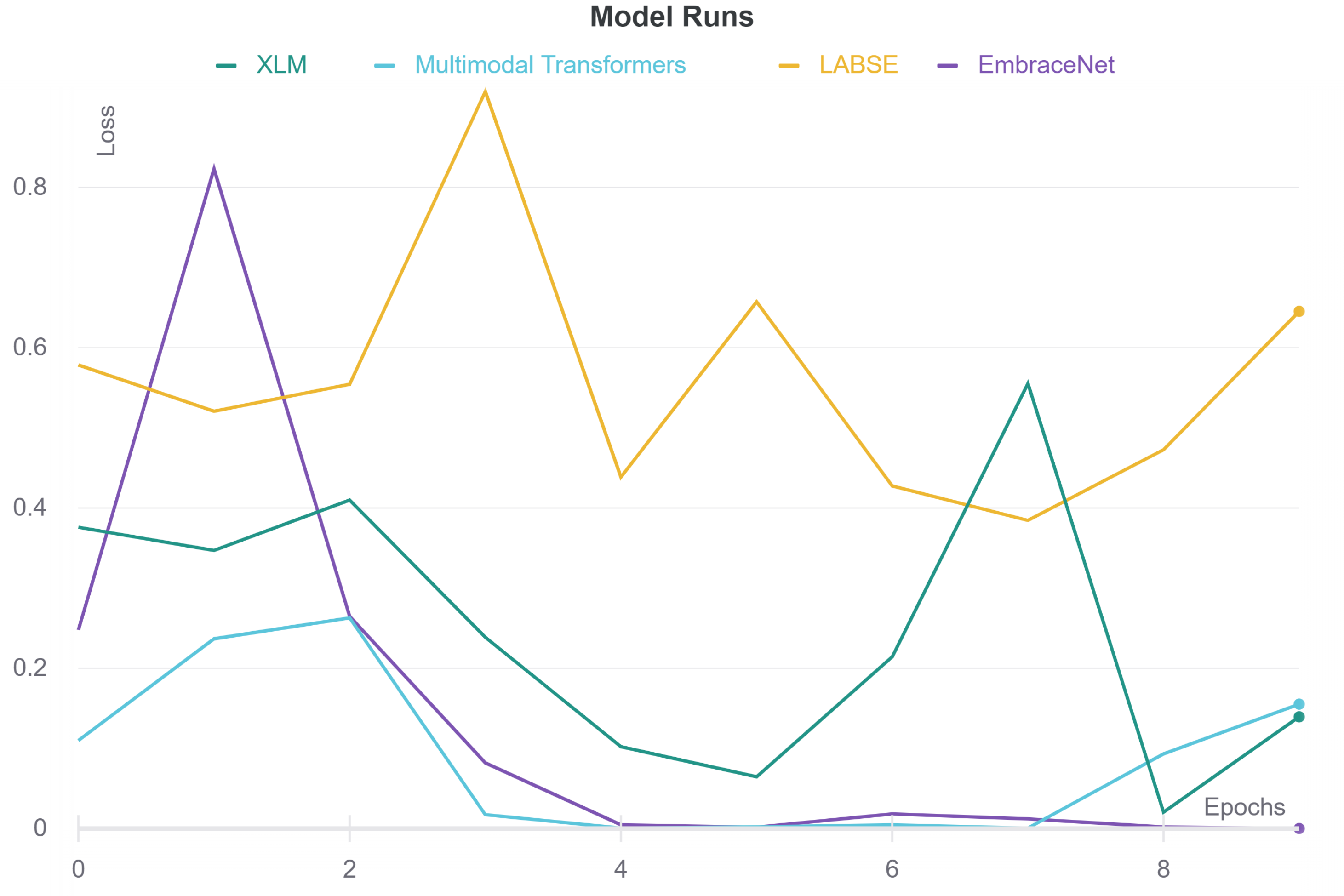}
    \caption{Losses during training}
    \label{fig:7}
\end{figure}
The use of images could have improved the result if there were valuable features in the images that could predict the probability. By the scores shown in Table 3, it can be seen that there is no significant increase in the metrics. The exact optimal learning rate with Adam optimiser was used. However, the highest score obtained was from the Multimodal Transformers with a weighted F1 score of 0.55. The transformers overfit the training and validation set, and the cross attention too created more variance on the training set. Although the images were centre cropped to remove the standard text presented at the top and bottom position, the test set had multiple images merged in them and text all over the meme. The merged text was a significant drawback as ImageNet models cannot extract features due to the difference in the size of kernels and the text present. They would be considered as just noise by the vision models. Even a sequential feature extractor like a vision transformer could not extract features from the embedded text on the image. Thus, images are an extra computation with no significant contribution to the model in this task. It is better to use only the text instead of combining both modalities for analysis. Keeping both computational power and performance as a deciding factor, textual analysis is a preferred solution for Tamil Meme Classification.

\section{Error Analysis}
\label{Section 8}
The multimodal approaches performed more inferior than the textual models. One of the reasons is that ImageNet trained features are not sufficient enough to detect troll classes. ImageNet or ResNet models are trained to classify general objects like humans, vehicles and food. These objects may be present in the meme dataset; however, the text is also embedded into it along with images. Training such models without fine-tuning creates high variance as the models are very complex and deep which could result in more over-fitting. Higher-level information needs to be captured from the images. However, one of the limitations of this analysis is the sample size of the dataset. The dataset contains only 1154 training images. By analysing the images present in the dataset, it has been observed that the emotion-related features are not contributing to detecting troll or not-troll classes. 
\begin{figure}
    \centering
    \includegraphics[width=\linewidth, height=6cm]{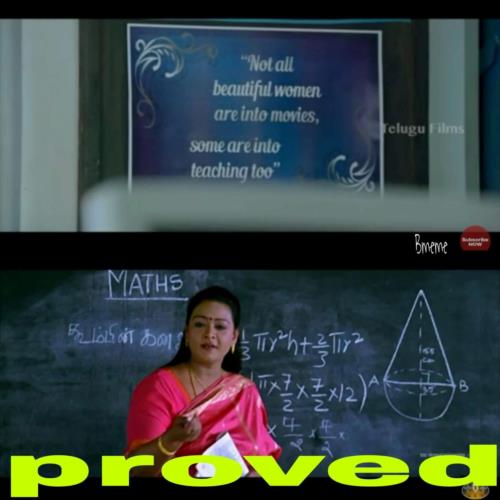}
    \caption{Example image in the test set belongs to troll class}
    \label{fig:8}
\end{figure}
\begin{figure}
    \centering
    \includegraphics[width=\linewidth, height=6cm]{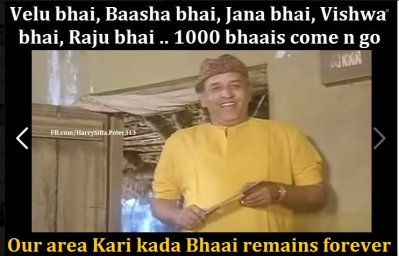}
    \caption{ Example image in the test set belongs to non-troll class}
    \label{fig:9}
\end{figure}
From Fig.\ref{fig:8} \& \ref{fig:9}, it has been noted that both images and facial expressions are almost the same however the classes are different. Another reason is that few images in the test set had only text embedded in all over the image which is shown in Fig.\ref{fig:9}
\begin{figure}
    \centering
    \includegraphics[width=\linewidth, height=6cm]{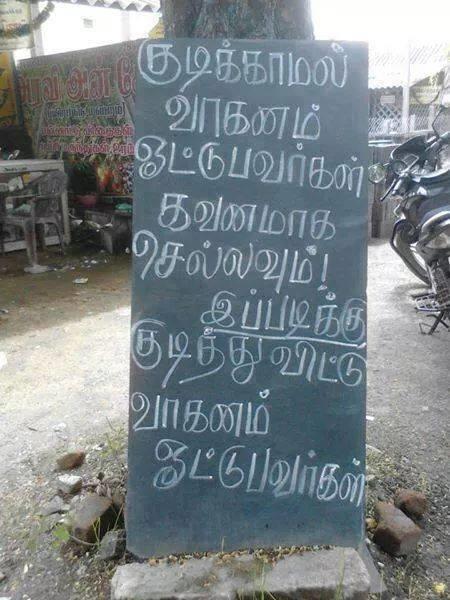}
    \caption{Example image with only the text embedded }
    \label{fig:10}
\end{figure}

\section{Conclusion}
This paper illustrates whether there is a need to include all the modalities for an experiment or not. It is not always true that features from every type of data will significantly contribute to predicting results. Multimodal analysis architectures from the simplest vanilla concatenation model to the most complex multimodal transformers with cross attention are shown. None of the multimodal models could exceed the textual model XLM with a weighted F1 score of 0.57. The distribution of the test set does matter, which has to be taken into consideration, and the type of images was different in the test set. That could mainly affect the performance of the ImageNet models while fine-tuning. In some cases, considering all modalities does create an immense increase in the result, where the computation power will not be wasted. In experiments like this task of meme classification with fewer data, it is always better to go for unimodal analysis rather than multimodal.   
\label{Section 9}
\begin{acknowledgements}
The author Bharathi Raja Chakravarthi was supported in part by a research grant from Science Foundation Ireland (SFI) under Grant Number SFI/12/RC/2289$\_$P2 (Insight$\_$2), co-funded by the European Regional Development Fund and Irish Research Council grant IRCLA/2017/129 (CARDAMOM-Comparative Deep Models of Language for Minority and Historical Languages).\\
\end{acknowledgements}
\section*{Funding}
This  research  has  not  been  funded  by  any  company  or organisation
\section*{Compliance with Ethical Standards}
\textbf{Conflict of interest:} The authors declare that they have no conflict of interest.\\
\\
\textbf{Availability of data and material:} The dataset used in this paper are obtained from \url{https://zenodo.org/record/4765573/}.\\
\\
\textbf{Code availability:} The data and approaches discussed in this paper are available at \url{https://github.com/adeepH/MemeClassification}.\\
\\
\textbf{Ethical Approval:}  This article does not contain any studies with human participants or animals performed by any of the authors.\\
 
 
%


%
%

\bibliographystyle{spmpsci}      
\bibliography{refs}   

%
%

\end{document}